\documentclass[10pt,twocolumn,letterpaper]{article}

\usepackage[pagenumbers]{wacv}      

\usepackage{times}
\usepackage{epsfig}
\usepackage{graphicx}
\usepackage{amsmath}
\usepackage{amssymb}
\usepackage{adjustbox}
\usepackage{multirow}
\usepackage{multicol}
\usepackage{booktabs}
\usepackage{url}
\usepackage{array}
\usepackage[dvipsnames,table,xcdraw]{xcolor}
\usepackage[accsupp]{axessibility}  
\usepackage{tikz}
\usetikzlibrary{fit}
\usepackage{pifont}

\definecolor{myorange}{rgb}{1, 0.647, 0}
\definecolor{myblue}{rgb}{.118, 0.565, 1}

\usepackage[pagebackref,breaklinks,colorlinks]{hyperref}

\usepackage[capitalize]{cleveref}
\crefname{section}{Sec.}{Secs.}
\Crefname{section}{Section}{Sections}
\Crefname{table}{Table}{Tables}
\crefname{table}{Tab.}{Tabs.}

\def\wacvPaperID{2181} 
\def\confName{WACV}
\def\confYear{2024}

\begin{document}

\title{Beyond Self-Attention: Deformable Large Kernel Attention for Medical Image Segmentation}

\author{Reza Azad$^{1}$
\quad Leon Niggemeier$^{1}$
\quad Michael Hüttemann$^{1}$ 
\quad Amirhossein Kazerouni$^{2}$\\
\quad Ehsan {Khodapanah Aghdam}$^{3}$ 
\quad Yury Velichko$^{4}$
\quad Ulas Bagci$^{4}$
\quad Dorit Merhof$\,$\footnotemark[1]$^{\;\,,5,6}$ 
\\
${^1}$ Faculty of Electrical Engineering and Information Technology, RWTH Aachen University, \\ Aachen, Germany\\
${^2}$ School of Electrical Engineering, Iran University of Science and Technology, Tehran, Iran\\
${^3}$ Department of Electrical Engineering, Shahid Beheshti University, Tehran, Iran \\
${^4}$ Department of Radiology, Northwestern University, Chicago, USA \\
${^5}$ Faculty of Informatics and Data Science, University of Regensburg, Regensburg, Germany\\
${^6}$ Fraunhofer Institute for Digital Medicine MEVIS, Bremen, Germany \\
{\tt\small }
}


\maketitle

\footnotetext[1]{Corresponding author. Email: \url{dorit.merhof@ur.de}.}
\thispagestyle{empty}

\begin{abstract}
Medical image segmentation has seen significant improvements with transformer models, which excel in grasping far-reaching contexts and global contextual information. However, the increasing computational demands of these models, proportional to the squared token count, limit their depth and resolution capabilities. Most current methods process D volumetric image data slice-by-slice (called pseudo 3D), missing crucial inter-slice information and thus reducing the model's overall performance.
To address these challenges, we introduce the concept of \textbf{Deformable Large Kernel Attention (D-LKA Attention)}, a streamlined attention mechanism employing large convolution kernels to fully appreciate volumetric context. This mechanism operates within a receptive field akin to self-attention while sidestepping the computational overhead. Additionally, our proposed attention mechanism benefits from deformable convolutions to flexibly warp the sampling grid, enabling the model to adapt appropriately to diverse data patterns. We designed both 2D and 3D adaptations of the D-LKA Attention, with the latter excelling in cross-depth data understanding. Together, these components shape our novel hierarchical Vision Transformer architecture, the \textit{D-LKA Net}. Evaluations of our model against leading methods on popular medical segmentation datasets (Synapse, NIH Pancreas, and Skin lesion) demonstrate its superior performance. Our code implementation is publicly available at the \href{https://github.com/mindflow-institue/deformableLKA}{GitHub}.

\end{abstract}

\section{Introduction}

Medical image segmentation plays a crucial role in computer-assisted diagnostics, aiding medical professionals in analyzing complex medical images. This process not only reduces the laboriousness of manual tasks and the dependence on medical expertise but also enables faster and more accurate diagnoses. The automation of segmentation offers the potential for faster and more accurate diagnostic outcomes, facilitating appropriate treatment strategies and enabling the execution of image-guided surgical procedures. Thus, the imperative to create rapid and precise segmentation algorithms serves as a driving force behind the motivation for this research.

Since the mid-2010s, Convolutional Neural Networks (CNNs) have become the preferred technique for many computer vision applications. Their ability to automatically extract complex feature representations from raw data without the need for manual feature engineering has generated significant interest within the medical image analysis community. Many successful CNN architectures such as U-Net \cite{ronneberger2015unet}, Fully Convolutional Networks \cite{long2015fully}, DeepLab \cite{chen2016semantic}, or SegCaps (segmentation capsules)~\cite{lalonde2021capsules} have been developed. These architectures have achieved great success in the task of semantic segmentation and have outperformed state-of-the-art (SOTA) methods previously \cite{azad2022medical,karimijafarbigloo2023mmcformer,karimijafarbigloo2023ms}.

The problem of identifying objects at different scales is a key concern in computer vision research \cite{lin2017feature,jose2023end}. In CNNs, the size of a detectable object is closely linked to the receptive field dimensions of the corresponding network layer. If an object extends beyond the boundaries of this receptive field, this may lead to under-segmentation outcomes. Conversely, using excessively large receptive fields compared to an object's actual size can limit recognition, as background information may exert undue influence on predictions \cite{He_2019_ICCV}.

A promising approach to address this issue involves employing multiple kernels with distinct sizes in parallel, similar to the mechanism of an \textit{Inception Block} \cite{szegedy2014going}. However, increasing the kernel sizes to accommodate larger objects is limited in practice due to the exponential increase in parameters and computational requirements \cite{He_2019_ICCV}. Consequently, various strategies, including pyramid pooling techniques \cite{he2015spatial} and dilated convolutions \cite{yu2015multi} at varying scales, have emerged to capture multi-scale contextual information.

Another intuitive concept entails directly incorporating multi-scale image pyramids or their associated feature representations into the network architecture. Yet, this approach poses challenges, particularly concerning the feasibility of managing training and inference times \cite{lin2017feature}. The use of Encoder-Decoder networks, such as U-Net, has proven advantageous in this context. Such networks encode appearance and location in shallower layers, while deeper layers capture higher semantic information and context from the broader receptive fields of neurons \cite{Lin_2018_ECCV}. Some methods combine features from different layers or predict features from layers of different sizes to use information from multiple scales~\cite{badrinarayanan2016segnet}. Also, forecasting features from layers of varied scales have emerged, effectively enabling the integration of insights across multiple scales \cite{lin2017feature}. However, most Encoder-Decoder structures face a challenge: they frequently fail to maintain consistent features across different scales and mainly use the last decoder layer to generate segmentation results \cite{azad2022medical,karimijafarbigloo2023self,He_2019_ICCV}.

Semantic segmentation is a task that involves predicting the semantic category for each pixel in an image based on a predefined label set. This task requires extracting high-level features while preserving the initial spatial resolution \cite{Mo_2022,Liu_2023}. CNNs are well suited to capture local details and low-level information, albeit at the expense of overlooking global context. This gap in handling global information has been a focus of the vision transformer (ViT) architecture, which has achieved remarkable success in several computer vision tasks, including semantic segmentation.

The cornerstone of the ViT is the \textit{attention mechanism}, which facilitates the aggregation of information across the entire input sequence. This capability empowers the network to incorporate long-range contextual cues beyond the limited receptive field sizes of CNNs \cite{Han_2023,shamshad2022transformers}. However, this strategy usually limits the ability of ViTs to effectively model local information \cite{azad2023advances}. 
This limitation can impede their ability to detect local textures, which is crucial for various diagnostic and prognostic tasks. This lack of local representation can be attributed to the particular way ViT models process images.
ViT models partition an image into a sequence of patches and model their dependencies using self-attention mechanisms. This approach may not be as effective as the convolution operations employed by CNN models for extracting local features within receptive fields. This difference in image processing methods between ViT and CNN models may explain the superior performance of CNN models in local feature extraction \cite{geirhosimagenet,azad2021texture}.
In recent years, innovative approaches have been developed to address the insufficient representation of local textures within Transformer models. One such approach involves integrating CNN and ViT features through complementary methods to combine their strengths and mitigate any local representation shortcomings \cite{chen2021transunet}. TransUNet \cite{chen2021transunet} is an early example of this approach, incorporating Transformer layers within the CNN bottleneck to model both local and global dependencies. HiFormer \cite{heidari2023hiformer} proposes a solution that combines a Swin Transformer module and a CNN-based encoder to generate two multi-scale feature representations that are integrated via a Double-Level Fusion module. UNETR \cite{hatamizadeh2022unetr} employs a Transformer-based encoder and a CNN decoder for 3D medical image segmentation. CoTr \cite{xie2021cotr} and TransBTS \cite{wang2021transbts} bridge the CNN encoder and decoder with the Transformer to enhance segmentation performance at low-resolution stages.

An alternate strategy to enhance local feature representation is to redesign the self-attention mechanism within pure Transformer models. In this vein, Swin-Unet \cite{cao2021swin} integrates a Swin Transformer \cite{liu2021swin} block of linear computational complexity within a U-shaped structure as a multi-scale backbone. MISSFormer \cite{huang2021missformer} employs the Efficient Transformer \cite{xie2021segformer} to address parameter issues in vision transformers by incorporating a non-invertible downsampling operation on input blocks. D-Former \cite{wu2022d} introduces a pure transformer-based pipeline featuring a double attention module to capture fine-grained local attention and interaction with diverse units in a dilated manner.
Nonetheless, certain specific limitations remain. These include computational inefficiency, as evidenced in the TransUNet model, the heavy dependence on a CNN backbone, as observed in HiFormer, and the lack of consideration of multi-scale information. Additionally, current segmentation architectures often take a slice-by-slice approach to process 3D input volumes, inadvertently disregarding potential correlations between neighboring slices. This oversight limits the comprehensive use of volumetric information, consequently compromising both localization accuracy and context integration. Furthermore, it is crucial to recognize that lesions in the medical domain often exhibit deformations in their shape. Therefore, any learning algorithm intended for medical image analysis must be endowed with the ability to capture and comprehend these deformations. Simultaneously, the algorithm should maintain computational efficiency to facilitate the processing of 3D volumetric data.

\textbf{Our contributions:} To address the challenges outlined above, we propose a solution in the form of the Deformable-LKA module (\ding{182}), which serves as a fundamental building block within our network design. This module is explicitly designed to effectively handle contextual information while simultaneously preserving local descriptors. This balance between the two aspects within our architecture enhances its ability to achieve precise semantic segmentation. Notably, our model introduces a dynamic adaptation of receptive fields based on the data, diverging from the conventional fixed filter masks found in traditional convolutional operations. This adaptive approach allows us to overcome the inherent limitations associated with static methods.
This innovative approach extends to the development of both 2D and 3D versions of the D-LKA Net architecture (\ding{183}). In the case of the 3D model, the D-LKA mechanism is tailored to suit a 3D context, thus enabling the seamless exchange of information across different volumetric slices.
(\ding{184}). Finally, our contribution is further emphasized by its computational efficiency. We achieve this through a design that leans solely on the D-LKA concept, resulting in a remarkable performance on various segmentation benchmarks that establish our method as a new SOTA approach.

\begin{figure*}[t]
    \centering
    \includegraphics[width=\textwidth]{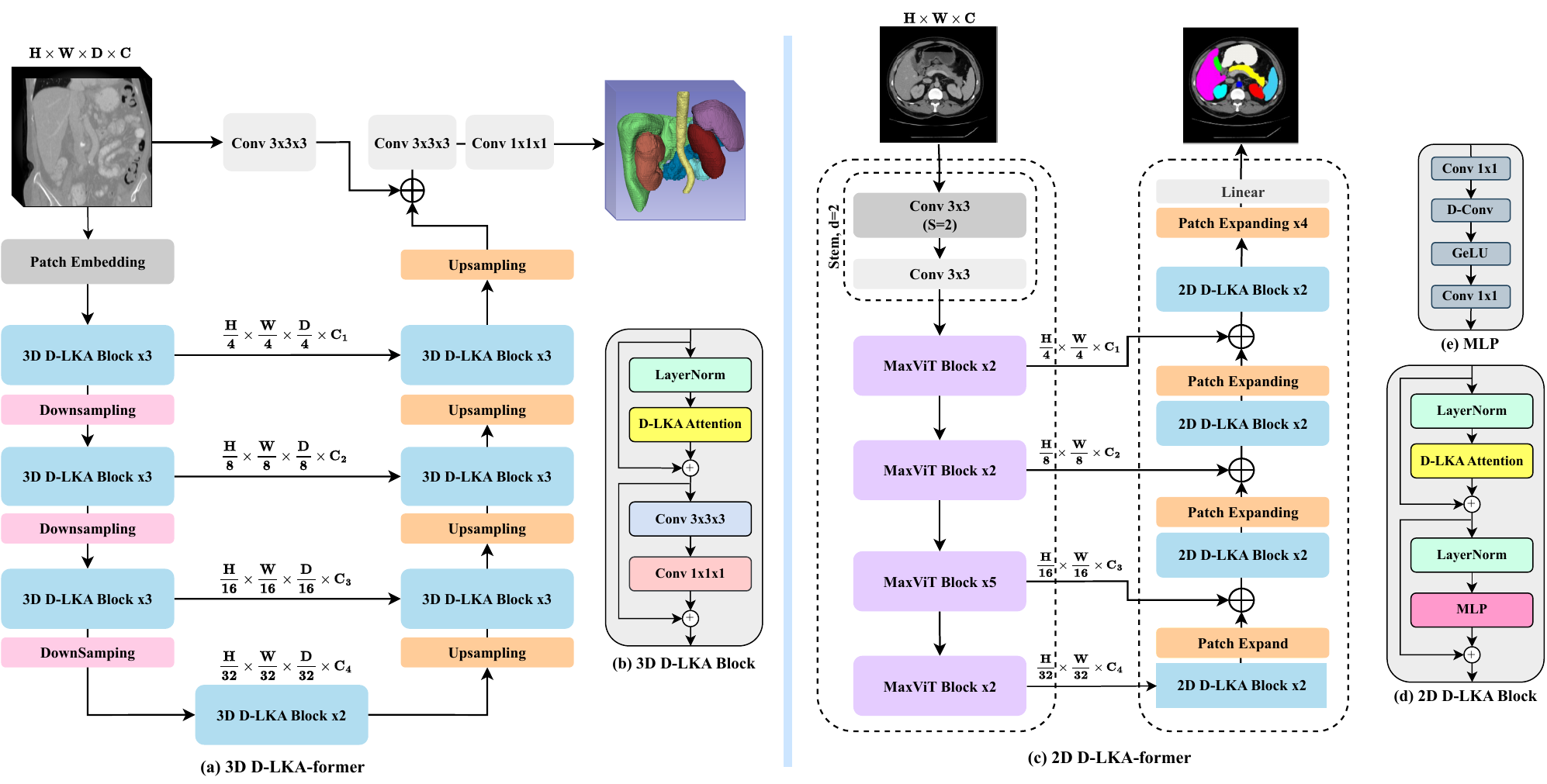}
    \caption{Proposed network architecture of the 3D D-LKA model on the left and the 2D D-LKA model on the right.}
    \label{fig:DLKA_Network}
\end{figure*}

\section{Method}
In this section, we begin by providing an overview of the methodology. First, we revisit the concept of Large Kernel Attention (LKA) as introduced by Guo et al. \cite{guo2022visual}. Then, we introduce our innovative exploration of the deformable LKA module. Built up on this, we introduce both 2D and 3D network architectures for segmentation tasks. 

\subsection{Large Kernel Attention}
Large convolution kernels provide a similar receptive field as the self-attention mechanism. A large convolution kernel can be constructed with much less parameters and computation by using a depth-wise convolution, a depth-wise dilated convolution, and a $1 \times 1$ convolution. The equations for the kernel sizes of the depthwise convolution and the depthwise dilated convolution to construct a $K \times K$ kernel for an input of dimension $H \times W$ and channels $C$ are:
\begin{equation}
    DW = \left(2d-1\right) \times \left(2d-1\right),
\end{equation}
\begin{equation}
    DW\text{-}D = \left\lceil\frac{K}{d}\right\rceil \times \left\lceil\frac{K}{d}\right\rceil,
\end{equation}
\noindent with a kernel size of $K$ and a dilation rate of $d$. 
The number of parameters $P\left(K,d\right)$ and floating-point operations (FLOPs) $F\left(K,d\right)$ is calculated by:
\begin{equation} \label{eq:lkaparams}
    P\left(K,d\right) = C\left(\left\lceil\frac{K}{d}\right\rceil^2  + \left(2d-1\right)^2 + 3 + C\right),
\end{equation}
\begin{equation}
    F\left(K,d\right) = P\left(K,d\right) \times H \times W.
\end{equation}
The number of FLOPs grows linearly with the size of the input image. The number of parameters increases quadratically with the number of channels and kernel size. However, as both are usually so small, they are not restricting factors.

To minimize the number of parameters for a fixed kernel size $K$, the derivative of equation \ref{eq:lkaparams} with respect to dilation rate $d$ can be set to zero:
\begin{equation}
    \frac{d}{d\hat{d}} P\left(K,\hat{d}\right)\overset{!}{=} 0 = C \cdot \left(8\hat{d}+ \frac{2K^2}{\hat{d}^3} - 4\right).
\end{equation}
For example, when the kernel size is $K=21$, this results in $d\approx 3.37$. Extending the formulas to the three-dimensional case is straightforward. For an input of size $H \times W \times D$ and channels $C$, then the equations for the number of parameters $P_{3d}\left(K,d\right) $ and FLOPs $F_{3d}\left(K,d\right)$ are:
\begin{equation}
    P_{3d}\left(K,d\right) = C\left(\left\lceil\frac{K}{d}\right\rceil^3  + \left(2d-1\right)^3 + 3 + C\right),
\end{equation}
\begin{equation}
    F_{3d}\left(K,d\right) = P_{3d}\left(K,d\right) \times H \times W\times D,
\end{equation}
\noindent with kernel size $K$ and dilation $d$.

\subsection{Deformable Large Kernel Attention} \label{sec:DeformLKA}
\begin{figure}[h]
    \centering
    \includegraphics[width=\columnwidth]{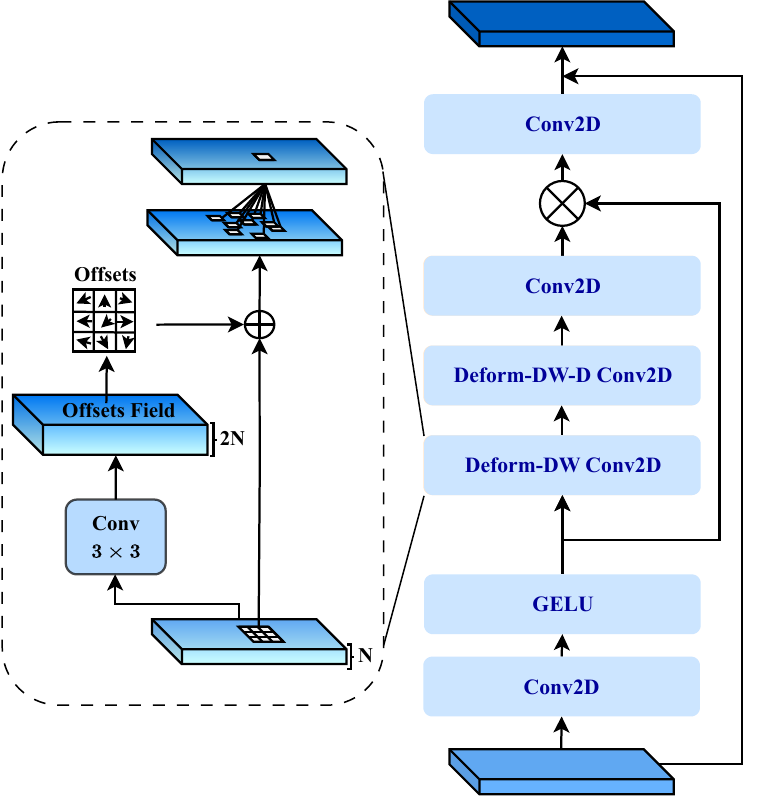}
    \caption{Architecture of the deformable LKA module.}
    \label{fig:D-LKA}
\end{figure}

The concept of utilizing Large Kernels for medical image segmentation is extended by incorporating Deformable Convolutions \cite{dai2017deformable}. Deformable Convolutions enable adjusting the sampling grid with whole-numbered offsets for free deformation. An additional convolutional layer learns The deformation from the feature maps, which creates an offset field. Learning the deformation based on the features itself results in an adaptive convolution kernel. 
This flexible kernel shape can improve the representation of lesions or organ deformations, resulting in an enhanced definition of object boundaries.
The convolutional layer responsible for calculating offsets follows the kernel size and dilation of its corresponding convolutional layer. Bilinear interpolation is employed to compute pixel values for offsets that are not found on the image grid. 
As shown in Figure~\ref{fig:D-LKA}, the D-LKA module can be formulated as:

\begin{equation}
\begin{aligned}
\text{Attention} &= \text{Conv}{1 \times 1}(\text{DDW}\text{-D-}\text{Conv}(\text{DDW}\text{-}\text{Conv}(\text{F}'))), \\
\text{Output} &= \text{Conv}{1 \times 1}(\text{Attention} \otimes \text{F}') + \text{F},
\end{aligned}
\end{equation}
\noindent where the input feature is denoted by $F \in \mathbb{R}^{C \times H \times W}$ and $F' = GELU(Conv(F))$. The attention component $\in$ $\mathbb{R}^{C \times H \times W}$ is represented as an attention map, with each value indicating the relative importance of corresponding features. The operator $\otimes$ denotes an element-wise product operation. Notably, LKA departs from conventional attention methods by not requiring additional normalization functions such as sigmoid or Softmax. According to \cite{wang2022antioversmoothing}, these normalization functions tend to neglect high-frequency information, thereby decreasing the performance of self-attention-based methods.

In the 2D version of the approach, the convolution layers are substituted with deformable convolutions because deformable convolutions improve the ability to capture objects characterized by irregular shapes and sizes. Such objects are commonly found in medical image data, making this augmentation especially significant.

However, extending the concept of deformable LKA to the 3D domain presents certain challenges. The primary constraint arises from the additional convolution layer needed for offset generation. In contrast to the 2D case, this layer cannot be executed in a depth-wise manner due to the nature of input and output channels. In the 3D context, input channels correspond to features, and the output channels enlarge to $3 \cdot k \times k \times k$, with a kernel size of $k$. The intricate nature of large kernels leads to an expansion of channel count along the third dimension, causing a substantial rise in parameters and FLOPs.
Consequently, an alternative approach is implemented for the 3D scenario. A sole deformable convolution layer is introduced into the existing LKA framework, following the depth-wise convolutions. This strategic design adaptation aims to mitigate the challenges posed by the extension to three dimensions.

\subsection{2D D-LKA Net}
The architecture of the 2D network is illustrated in Figure \ref{fig:DLKA_Network}. The first variant uses a MaxViT \cite{tu2022maxvit} as the encoder component for efficient feature extraction, while the second variant incorporates deformable LKA layers for more refined, superior segmentation.

In a more formal description, the encoder generates four hierarchical output representations. A convolutional stem first reduces the dimensions of the input image to $ \frac{H}{4} \times \frac{W}{4} \times C$. Subsequently, feature extraction is carried out through four stages of MaxViT blocks, each followed by downsampling layers. As the process advances to the decoder, four stages of D-LKA layers are implemented, each stage encompassing two D-LKA blocks. Patch-expanding layers are then applied to achieve resolution upsampling while also decreasing channel dimensions. Finally, a linear layer is responsible for generating the ultimate output.

The structure of the 2D D-LKA block comprises LayerNorm, deformable LKA, and a Multi-Layer Perceptron (MLP). The integration of residual connections ensures effective feature propagation, even across deeper layers. This arrangement can be mathematically represented as:
\begin{equation}
    x_1 = D\text{-}LKA\text{-}Attn(LN(x_{in})) + x_{in},
\end{equation}
\begin{equation}
    x_{out} = MLP(LN(x_1)) + x_1,
\end{equation}
\begin{equation}
    MLP = Conv_1(GeLU(Conv_d(Conv_1(x)))),
\end{equation}
\noindent with input features $x_{in}$, layer normalization $LN$, deformable LKA attention $D\text{-}LKA\text{-}Attn$, depthwise convolution $Conv_d$, linear layers $Conv_1$ and GeLU activation function $GeLU$.

\subsection{3D D-LKA Net}
The 3D network architecture, depicted in Figure \ref{fig:DLKA_Network}, is structured hierarchically using an encoder-decoder design. Initially, a patch embedding layer reduces the input image dimensions from $(H \times W \times D)$ to $(\frac{H}{4} \times \frac{W}{4} \times \frac{D}{2})$. Within the encoder, a sequence of three D-LKA stages is employed, each containing three D-LKA blocks. After each stage, a downsampling step reduces spatial resolution by half while doubling the channel dimension. The central bottleneck includes another set of two D-LKA blocks. The decoder structure is symmetric to that of the encoder. In order to double the feature resolution while reducing channel count, transpose convolutions are utilized. Each decoder stage employs three D-LKA blocks to promote long-range feature dependencies. The final segmentation output is produced by a $3 \times 3 \times 3$ convolutional layer, succeeded by a $1 \times 1 \times 1$ convolutional layer to match class-specific channel requirements. To establish a direct connection between the input image and the segmentation output, a skip connection is formed using convolutions. Additional skip connections perform a fusion of features from other stages based on simple addition. The ultimate segmentation map is produced through a combination of $3 \times 3 \times 3$ and $1 \times 1 \times 1$ convolutional layers.

The 3D D-LKA block includes layer normalization followed by D-LKA Attention, with residual connections applied. The subsequent section employs a $3 \times 3 \times 3$ convolutional layer, followed by a $1 \times 1 \times 1$ convolutional layer, both accompanied by residual connections. This entire process can be summarized as follows:
\begin{equation}
    x_1 = DAttn(LN(x_{in})) + x_{in},
\end{equation}
\begin{equation}
    x_{out} = Conv_1(Conv_3(x_1)) + x_1,
\end{equation}
\noindent with input features $x_{in}$, layer normalization $LN$, deformable LKA $DAttn$, convolutional layer $Conv_1$, and output features $x_{out}$. $Conv_3$ refers to a feed forward network, with two convolutional layers and activation functions.

\section{Experiments}
\subsection{Experimental Setup}
We have implemented both 2D and 3D models using the PyTorch framework and performed training on a single RTX 3090 GPU. For the 2D method, a batch size of 20 was used, along with Stochastic Gradient Descent (SGD) employing a base learning rate of 0.05, a momentum of 0.9, and a weight decay of 0.0001. The training process consisted of 400 epochs, employing a combination of cross-entropy and Dice loss, represented as follows:

\begin{equation}
\mathcal{L}_{total} = 0.6 \cdot \mathcal{L}_{dice} + 0.4 \cdot \mathcal{L}_{ce}.
\end{equation}

Consistent with \cite{chen2021transunet}, identical data augmentation techniques were applied.
For the 3D model, a batch size of 2 was chosen, and stochastic gradient descent was employed with a base learning rate of 0.01 and a weight decay of $3e^{-5}$. The input images were in the form of patches sized $128 \times 128 \times 64$. The training process consisted of 1000 epochs, with 250 patches utilized per epoch. Data augmentation techniques consistent with nnFormer \cite{zhou2021nnformer} and UNETR++ \cite{shaker2022unetr++} were employed.

\subsection{Datasets}
\noindent\textbf{Synapse Multi-Organ Segmentation}:
First, we evaluate the performance of our method using the well-established synapse multi-organ segmentation dataset \cite{synapse2015ct}. This dataset consists of 30 cases with a total of 3779 axial abdominal clinical CT images. Each CT volume comprises a range of $85$ to $198$ slices, each with dimensions of $512 \times 512$ pixels. The voxel spatial resolution varies in the range of $([0.54 \sim 0.54] \times[0.98 \sim 0.98] \times[2.5 \sim 5.0])$ $\mathrm{mm}^{3}$. Our evaluation follows the setting presented in \cite{chen2021transunet,shaker2022unetr++} for 2D and 3D versions.

\noindent\textbf{Skin Lesion Segmentation}:
Our comprehensive experiments also extend to skin lesion segmentation datasets. Specifically, we leverage the ISIC 2017 dataset \cite{codella2018skin}, which contains 2000 dermoscopic images for training, 150 for validation, and 600 for testing. Furthermore, we adopt the division scheme described in previous literature \cite{asadi2020multi,azad2019bi,aghdam2022attention,eskandari2023inter} for the ISIC 2018 dataset \cite{codella2019skin}. Additionally, the $\mathrm{PH}^{2}$ dataset \cite{mendoncca2013ph} is used as a dermoscopic image repository designed for both segmentation and classification tasks. This dataset consists of 200 dermoscopic images containing 160 nevi and 40 melanomas.

\noindent\textbf{NIH Pancreas Segmentation}:
The publicly available NIH Pancreas dataset consists of 82 contrast-enhanced 3D abdominal CT volumes, each accompanied by manual annotations \cite{roth2015deeporgan}. In our configuration, we use 62 samples for training and reserve the remaining samples for testing.

\subsection{Quantitative and Qualitative Results}
\begin{table*}[!ht]
    \begin{center}
    \caption{Comparison results of the proposed method evaluated on the Synapse dataset.  \textcolor{blue}{Blue} indicates the best result, and \textcolor{red}{red} displays the second-best. Parameters are reported in millions (M) and FLOPS in billions (G). DSC is presented for abdominal organs spleen (Spl), right kidney (RKid), left kidney (LKid), gallbladder (Gal), liver (Liv), stomach (Sto), aorta (Aor), and pancreas (Pan).}  
        \resizebox{\textwidth}{!}{
        \begin{tabular}{l|c c| c c c c c c c c|cc}
        \toprule
            \multirow{2}{*}{Methods} & \multirow{2}{*}{Params (M)} & \multirow{2}{*}{FLOPs (G)} & \multirow{2}{*}{Spl} &  \multirow{2}{*}{RKid} &  \multirow{2}{*}{ LKid} & \multirow{2}{*}{Gal}  & \multirow{2}{*}{Liv}  & \multirow{2}{*}{Sto} & \multirow{2}{*}{Aor} &  \multirow{2}{*}{Pan} &  \multicolumn{2}{c}{Average} 
            
            \\ \cmidrule{12-13}
             & & & & & & & & & & & DSC $\uparrow$ & HD95 $\downarrow$ \\
                \midrule
                \midrule
        
        TransUNet \cite{chen2021transunet}~ & 96.07 & 88.91 & 85.08 & 77.02 & 81.87 & 63.16 & 94.08 & 75.62 &  87.23  & 55.86 & 77.49 & 31.69 \\
        Swin-UNet \cite{cao2021swinunet}~ & 27.17 &6.16 & 90.66 &  79.61 & 83.28 & 66.53 & 94.29 & 76.60 & 85.47 &   56.58  & 79.13 & 21.55 \\
        LeViT-UNet-384 \cite{xu2021levit}~ & 52.17 & 25.55 & 88.86 & 80.25 & 84.61 & 62.23 & 93.11 & 72.76 & 87.33 & 59.07 & 78.53 & 16.84\\
        MISSFormer \cite{huang2022missformer}~ & 42.46 & 9.89 & \textcolor{blue}{91.92} &  82.00 & 85.21 &  68.65 & 94.41 & 80.81 & 86.99 & 65.67 & 81.96 & 18.20 \\
        ScaleFormer \cite{huang2022scaleformer}~ & 111.6 & 48.93 & 89.40 &  83.31 & 86.36 &  \textcolor{blue}{74.97} & \textcolor{blue}{95.12} & 80.14 & \textcolor{blue}{88.73} & 64.85 & \textcolor{red}{82.86} & 16.81 \\
        HiFormer-B\cite{heidari2022hiformer}~ &25.51 & 8.045 & 90.99 &  79.77 & 85.23 &  65.23 & 94.61 & 81.08 & 86.21 & 59.52 & 80.39 & \textcolor{blue}{14.70} \\
        DAEFormer \cite{azad2022dae}~ & 48.07 & 27.89 & \textcolor{red}{91.82} &  \textcolor{red}{82.39} & \textcolor{red}{87.66} &  71.65 & \textcolor{red}{95.08} & 80.77 & 87.84 & 63.93 & 82.63 & \textcolor{red}{16.39} \\
        TransDeepLab \cite{azad2022transdeeplab}~ & 21.14 & 16.31 & 89.00 &  79.88 & 84.08 &  69.16 & 93.53 & 78.40 & 86.04 & 61.19 & 80.16 & 21.25 \\
        PVT-CASCADE \cite{rahman2023medical}~ & 35.28 & 6.40 & 90.1 &  80.37 & 82.23 &  70.59 & 94.08 & 83.69 & 83.01 & 64.43 & 81.06 & 20.23 \\
        \rowcolor[HTML]{C8FFFD}
        \midrule
        LKA Baseline & 85.82 & 13.62 &  91.45 & 81.93 & 84.93 & 71.05 & 94.87 & \textcolor{red}{83.71} & 87.48 & \textcolor{red}{66.76} & 82.77 & 17.42 \\
        \rowcolor[HTML]{C8FFFD}
        \textbf{2D D-LKA Net}~ & 101.64 &	19.92 & 91.22 & \textcolor{blue}{84.92} & \textcolor{blue}{88.38} & \textcolor{red}{73.79} & 94.88 & \textcolor{blue}{84.94} & \textcolor{red}{88.34} & \textcolor{blue}{67.71} & \textcolor{blue}{84.27} & 20.04 \\
        \midrule
        \bottomrule
        \end{tabular}
        }\vspace{0.5em}
        
        \label{table:2D_synapse_results}
        \end{center}
\vspace{-0.8cm}
\end{table*}

\subsubsection{2D results}
\textbf{Synapse Dataset:} 
In Table \ref{table:2D_synapse_results}, we present a comprehensive comparison of the leading performances achieved by other SOTA techniques in contrast to our proposed approach.

\begin{figure}[h]
    \centering
    \includegraphics[width=\columnwidth]{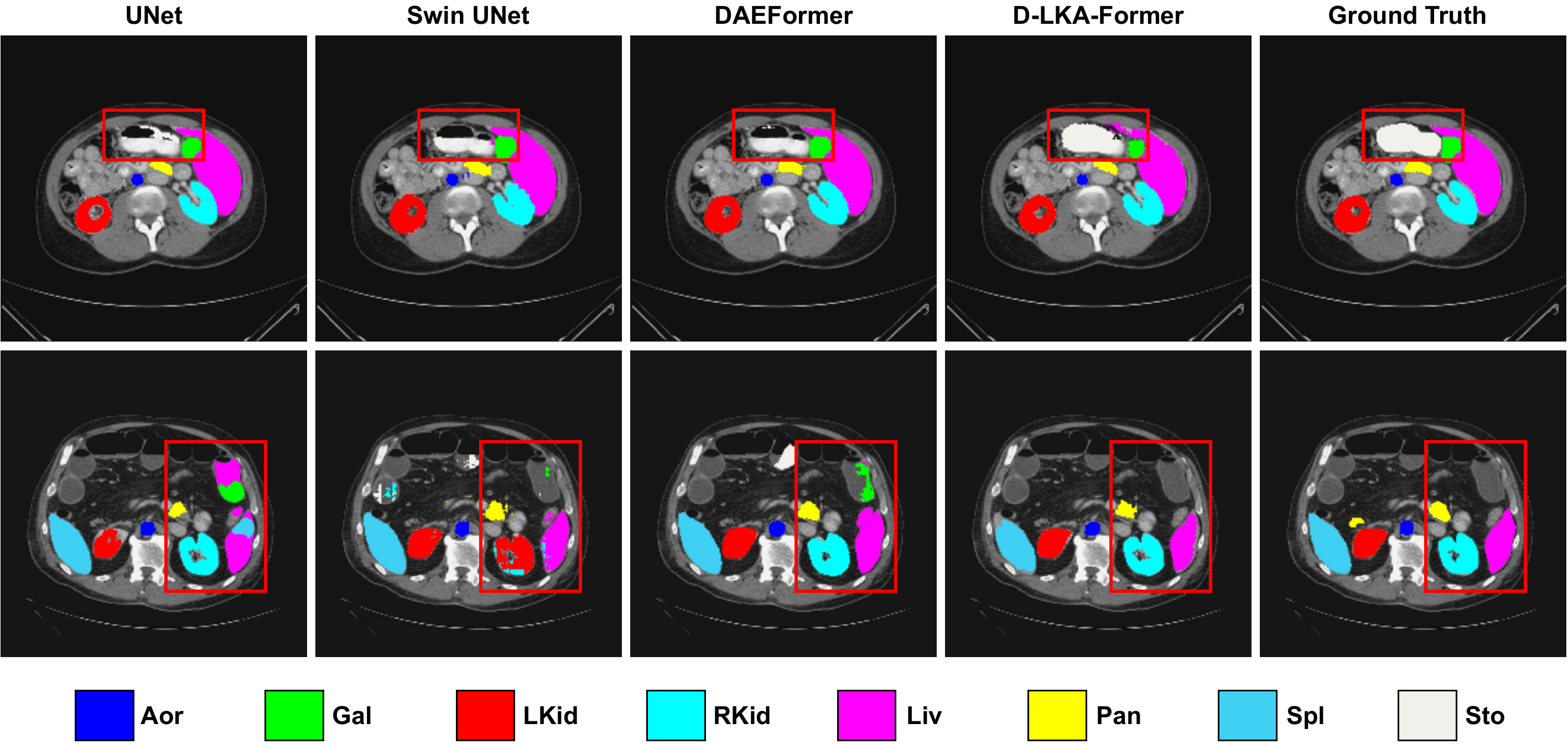}
    \caption{Qualitative comparison of the results from UNet \cite{ronneberger2015unet}, Swin UNet \cite{cao2022swin}, DAEFormer \cite{azad2022dae} and D-LKA Net.}
    \label{fig:2d_visualization}
\end{figure}

The results in terms of the Dice Similarity Coefficient (DSC) reveal that D-LKA Net exhibits superiority over the previously established SOTA methods. Specifically, it outperforms ScaleFormer \cite{huang2022scaleformer} by 1.41\%, DAEFormer \cite{azad2022dae} by 1.64\%, and by even larger margins when compared to other approaches. Notably, significant improvements are seen in the segmentation of specific anatomical regions such as the right kidney, left kidney, stomach, and pancreas. Notably, the pancreas achieves a significantly improved segmentation result, showing an impressive 2.04\% improvement over the second-best performing method. Given that the segmentation of smaller abdominal organs, such as the gallbladder or pancreas, has historically been a challenge for existing SOTA approaches, this notable performance improvement represents a significant step forward in achieving more accurate segmentation results.
A qualitative comparison of different methods is shown in Figure \ref{fig:2d_visualization}. Compared to DAE-Former \cite{azad2022dae}, our approach results in less misclassifications for the \textit{stomach}. While Unet \cite{ronneberger2015unet} and Swin-UNet \cite{cao2022swin} sometimes classify distant tissue as parts of the \textit{liver, gallbladder} or \textit{stomach}, our approach reduces the misclassifications and better represents the shape of the organs.

\noindent\textbf{Skin Lesion Segmentation Results:} 
The comparative outcomes for skin lesion segmentation benchmarks, including ISIC 2017, ISIC 2018, and $\mathrm{PH}^{2}$, in contrast to leading methods, are detailed in Table~\ref{tab:skin_comparison}. Notably, our D-LKA Net consistently outperforms its competitors across various evaluation metrics. This consistent superiority observed across different datasets underscores D-LKA Net's robust generalization capabilities.

\begin{table*}[!ht]
    \centering
    \caption{Performance comparison of the proposed method against the SOTA approaches on skin lesion segmentation benchmarks.}
    \resizebox{\textwidth}{!}{
    \begin{tabular}{l|cccc|cccc|cccc} 
    \toprule
    \multirow{2}{*}{\textbf{Methods}} & \multicolumn{4}{c|}{\textbf{ISIC 2017}} &  \multicolumn{4}{c|}{\textbf{ISIC 2018}}  & \multicolumn{4}{c}{$\mathbf{PH^2}$} \\ 
    \cline{2-13}
     & \textbf{DSC} & \textbf{SE} & \textbf{SP} & \textbf{ACC} & \textbf{DSC} & \textbf{SE} & \textbf{SP} & \textbf{ACC} & \textbf{DSC} & \textbf{SE} & \textbf{SP} & \textbf{ACC} \\ 
    \midrule
    U-Net \cite{ronneberger2015unet} & 0.8159 & 0.8172 & 0.9680 & 0.9164 & 0.8545 & 0.8800 & 0.9697 & 0.9404 & 0.8936 & 0.9125  & 0.9588  & 0.9233 
    \\
    Att-UNet \cite{schlemper2019attention} & 0.8082 & 0.7998 & 0.9776 & 0.9145 & 0.8566 & 0.8674 & \textcolor{blue}{0.9863} & 0.9376 & 0.9003 & 0.9205  & 0.9640  & 0.9276 
    \\
    DAGAN \cite{lei2020skin} & 0.8425 & 0.8363 & 0.9716 & 0.9304 & 0.8807 & 0.9072 & 0.9588 & 0.9324 & 0.9201 & 0.8320  & 0.9640  & 0.9425 
    \\
    TransUNet \cite{chen2021transunet} & 0.8123 & 0.8263 & 0.9577 & 0.9207 & 0.8499 & 0.8578 & 0.9653 & 0.9452 & 0.8840 & 0.9063  & 0.9427  & 0.9200 
    \\
    MCGU-Net \cite{asadi2020multi} & 0.8927 & 0.8502 & \textcolor{red}{0.9855} & 0.9570 & 0.8950 & 0.8480 & \textcolor{red}{0.9860}  & 0.9550 & 0.9263 & 0.8322  & 0.9714  & 0.9537 
    \\
    MedT \cite{valanarasu2021medical} & 0.8037 & 0.8064 & 0.9546 & 0.9090 & 0.8389 & 0.8252 & 0.9637 & 0.9358 & 0.9122 & 0.8472  & 0.9657  & 0.9416 
    \\
    FAT-Net \cite{wu2022fat} & 0.8500 & 0.8392 & 0.9725 & 0.9326 & 0.8903 & 0.9100 & 0.9699 & 0.9578 & 0.9440 & \textcolor{blue}{0.9441}  & 0.9741  & \textcolor{blue}{0.9703} 
    \\
    TMU-Net \cite{azad2022contextual} & 0.9164 & 0.9128 & 0.9789 & 0.9660 & 0.9059 & 0.9038 & 0.9746 & 0.9603 & 0.9414 & 0.9395  & 0.9756 & 0.9647 
    \\
    Swin-Unet \cite{cao2021swinunet} & 0.9183 & 0.9142 & 0.9798 & 0.9701 & 0.8946 & 0.9056 & 0.9798 & \textcolor{red}{0.9645} & 0.9449 & 0.9410 & 0.9564 & \textcolor{red}{0.9678} 
    \\
    DeepLabv3+ (CNN) \cite{chen2018encoder} & 0.9162 & 0.8733 & \textcolor{blue}{0.9921} & 0.9691 & 0.8820 & 0.8560 & 0.9770 & 0.9510 & 0.9202 & 0.8818 & \textcolor{blue}{0.9832} & 0.9503 
    \\
    HiFormer-B \cite{heidari2023hiformer} & \textcolor{red}{0.9253} & 0.9155 & 0.9840 & 0.9702 & 0.9102 & \textcolor{red}{0.9119} & 0.9755 & 0.9621 & \textcolor{red}{0.9460} & 0.9420 & 0.9772 & 0.9661 
    \\
    \midrule
    \rowcolor[HTML]{C8FFFD}
    LKA (Baseline) & 0.9229 & \textcolor{red}{0.9192} & 0.9824 & \textcolor{red}{0.9704} & \textcolor{red}{0.9118} & 0.8984 & 0.9802 & 0.9632 & 0.9409 & 0.9312 & 0.9759 & 0.9612 \\
    \rowcolor[HTML]{C8FFFD}
    \textbf{2D D-LKA Net} & \textcolor{blue}{0.9254} & \textcolor{blue}{0.9327} & 0.9793 & \textcolor{blue}{0.9705} & \textcolor{blue}{0.9177} & \textcolor{blue}{0.9164} & 0.9773 & \textcolor{blue}{0.9647} & \textcolor{blue}{0.9490} & \textcolor{red}{0.9430} & \textcolor{red}{0.9775} & 0.9659\\
    \bottomrule
    \end{tabular}}
    
    \label{tab:skin_comparison}
\end{table*}

\begin{figure}[htbp]
    \centering
    \includegraphics[width=\columnwidth]{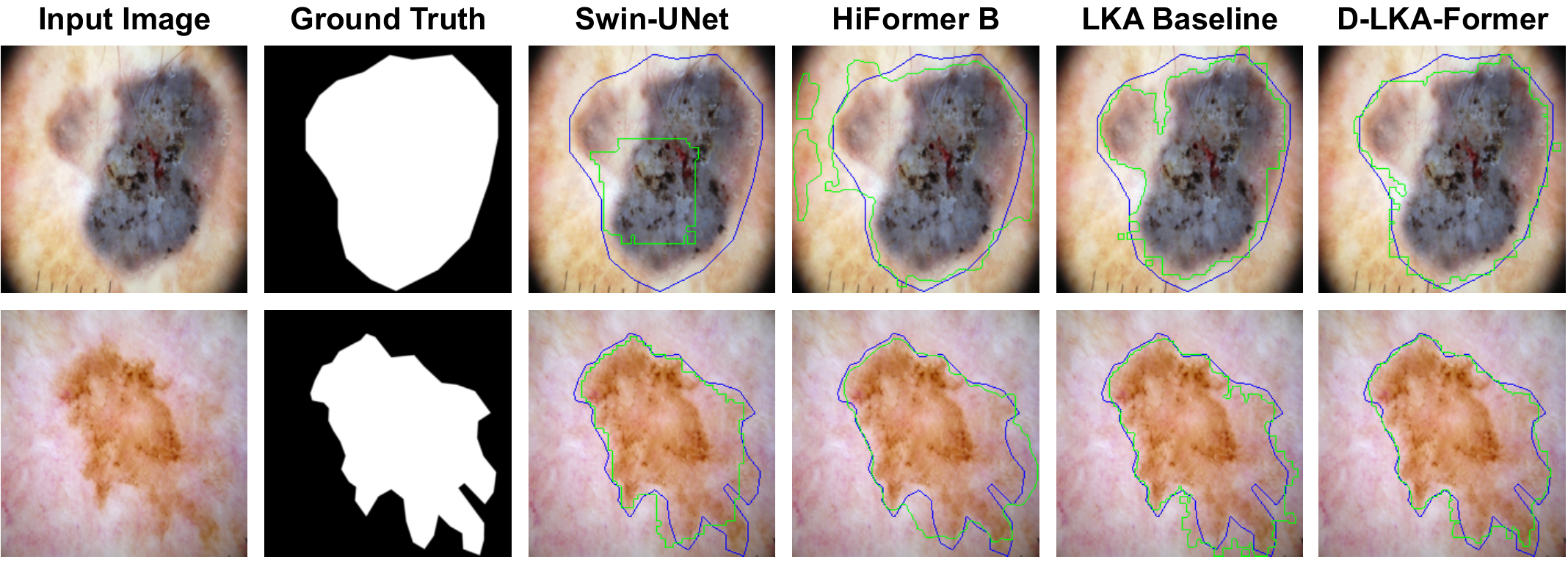}
    \caption[2D Skin visualizations]{2D visualizations on the ISIC 2018 \cite{codella2019skin} dataset.}
	\label{fig:2D_skin_results_visualization}
\end{figure}

A qualitative comparison of the results is presented in Figure \ref{fig:2D_skin_results_visualization}. In comparison to the baseline method, D-LKA Net better follows the complex outline of the lesions. In contrast to Swin-UNet and HiFormer-B, which tend to over- or under-segment certain areas, our approach achieves a more accurate segmentation.\\

\subsubsection{3D results}
\begin{figure*}[ht]
    \centering
    \includegraphics[width=\textwidth]{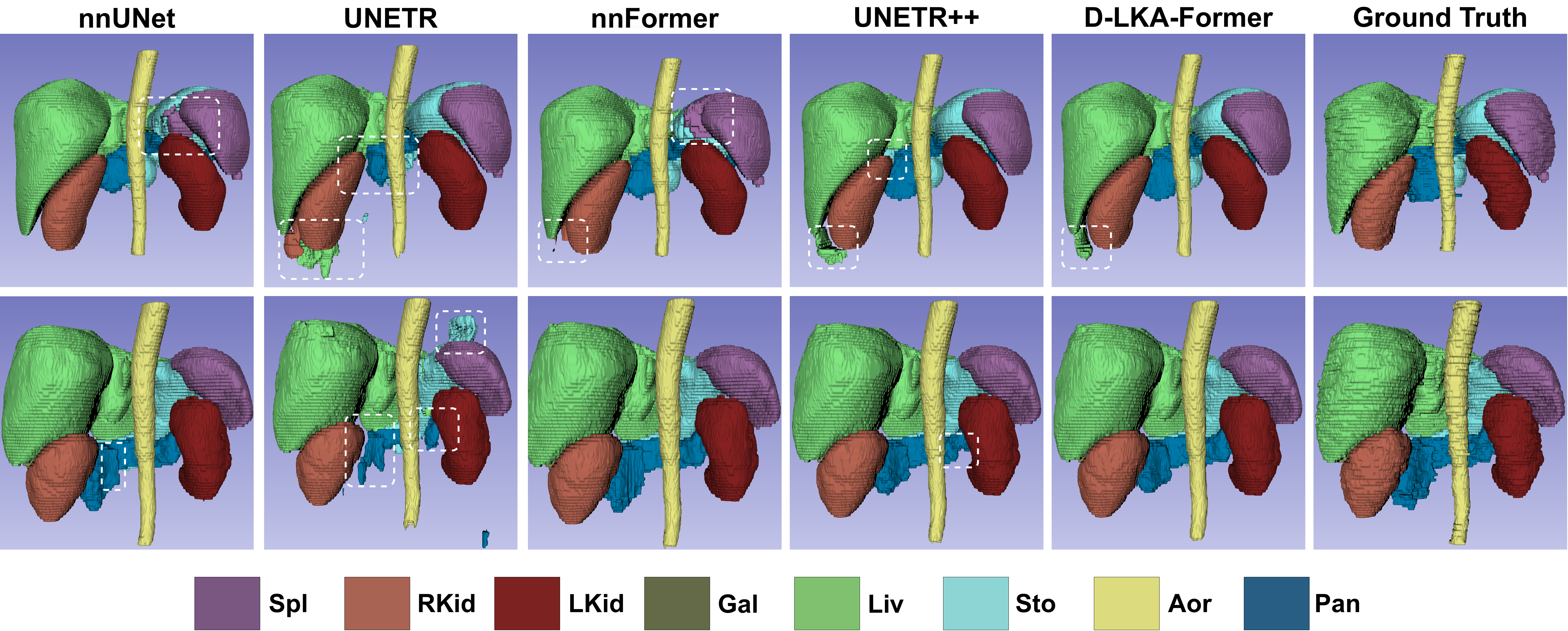}
    \caption{Qualitative results on the Synapse dataset. The boxes mark regions of failure.}
    \label{fig:synapse_3d_visual}
\end{figure*}

\textbf{Synapse Dataset:} We compare our 3D approach with previous SOTA methods on the Synapse dataset. The results are presented in Table \ref{tab:3D_synapse_results}.
\begin{table*}[!ht]
    \begin{center}
    \caption{Results on the Synapse dataset using 3D models. \textcolor{blue}{Blue} indicates the best result, and \textcolor{red}{red} displays the second-best.}
        \resizebox{\textwidth}{!}{
        \begin{tabular}{l|c c| c c c c c c c c|cc}
        \toprule
            \multirow{2}{*}{Methods} & \multirow{2}{*}{Params (M)} & \multirow{2}{*}{FLOPs (G)} & \multirow{2}{*}{Spl} &  \multirow{2}{*}{RKid} &  \multirow{2}{*}{ LKid} & \multirow{2}{*}{Gal}  & \multirow{2}{*}{Liv}  & \multirow{2}{*}{Sto} & \multirow{2}{*}{Aor} &  \multirow{2}{*}{Pan} &  \multicolumn{2}{c}{Average} 
            
            \\ \cmidrule{12-13}
             & & & & & & & & & & & DSC $\uparrow$ & HD95 $\downarrow$ \\
            \midrule
            \midrule
        UNETR \cite{hatamizadeh2022unetr} & 92.49 & 75.76 & 85.00 & 84.52 & 85.60 & 56.30 & 94.57 & 70.46 & 89.80 & 60.47 & 78.35 & 18.59 \\
        Swin-UNETR \cite{hatamizadeh2021swinunetr} & 62.83 & 384.2 & 95.37 & 86.26 & 86.99 & 66.54 & 95.72 & 77.01 & 91.12 & 68.80 & 83.48 & 10.55 \\
        nnFormer \cite{zhou2021nnformer} & 150.5 & 213.4 & 90.51 & 86.25 & 86.57 & 70.17 & \textcolor{red}{96.84} & \textcolor{blue}{86.83} & 92.04 & \textcolor{blue}{83.35} & 86.57 & 10.63 \\
        UNETR++ \cite{shaker2022unetr++} & 42.96 & 47.98 & \textcolor{red}{95.77} & 87.18 & 87.54 & \textcolor{red}{71.25} & 96.42 & \textcolor{red}{86.01} & 92.52 & 81.10 & \textcolor{red}{87.22} & \textcolor{blue}{7.53}\\
        \midrule
        \rowcolor[HTML]{C8FFFD}
        LKA Baseline & 28.94 & 48.79 & 90.49 & \textcolor{red}{87.54} & \textcolor{red}{87.57} & 63.81 & \textcolor{blue}{96.96} & 84.89 & \textcolor{blue}{93.22} & \textcolor{red}{82.46} & 85.87 & 14.35 \\
        \rowcolor[HTML]{C8FFFD}
        \textbf{D-LKA Net} & 42.35 & 66.96 & \textcolor{blue}{95.88} & \textcolor{blue}{88.50} & \textcolor{blue}{87.64} & \textcolor{blue}{72.14} & 96.25 & 85.03 & \textcolor{red}{92.87} & 81.64 & \textcolor{blue}{87.49} & \textcolor{red}{9.57} \\
        \midrule
        \bottomrule
        \end{tabular}
        }\vspace{0.5em}

        \label{tab:3D_synapse_results}
        \end{center}
\vspace{-0.8cm}
\end{table*}
We achieve a 0.27\%  improvement in \textit{DSC} over the previous SOTA methods UNETR++ \cite{shaker2022unetr++}. Compared to nnFormer \cite{zhou2021nnformer}, an improvement of 0.92\% is achieved. For the HD95 metric, D-LKA Net reaches the second-best result. In comparison to UNETR++, a small performance increase is observed for \textit{spleen, left kidney} and \textit{aorta}. A significant increase is reported of \textit{right kidney} and the small organs \textit{gallbladder} and \textit{pancreas}. An increase in the segmentation performance of these small organs is especially important.\\
In terms of parameters, we have the lowest number of parameters with only 42.35 M while still achieving excellent segmentation performance. The number of FLOPs is 66.96 G, the second lowest. Only UNETR++ has less FLOPs. Compared to SOTA approaches like Swin-UNETR and nnFormer, we need only about 17\% and 31\% of the computations while achieving a better performance. \\
Figure \ref{fig:synapse_3d_visual} shows qualitative results on the Synapse dataset. In comparison to the nnFormer, we better capture the \textit{aorta} as a whole and don't confuse other tissue as the organ. In comparison to UNETR++, we better segment the \textit{pancreas}, whereas UNETR++ tends to under-segment. Also, our approach segments the \textit{liver} and \textit{stomach} more accurately than UNETR++, which tends to over-segment these organs.

\noindent\textbf{Pancreas Dataset: } The results of the NIH Pancreas dataset are presented in Table \ref{tab:pancreas_results}. Our approach achieves the best performance in all four metrics. In comparison to the closest competitor UNETR++, an increase of 0.63\% in DSC, 0.82\% in Jaccard, and a decrease of 1.04 in HD95 and 0.26 in ASD can be noted. D-LKA Net also has the lowest number of Parameters, with 62.07 M.
\begin{table}[!ht]
    \begin{center}
    \caption[3D Pancreas results]{Results on the Pancreas dataset. The dice score, Jaccard index, HD95, and ASD are reported.}
        \resizebox{\columnwidth}{!}{
        \begin{tabular}{l|c c| c c c c}
        \toprule
            Methods & Params (M) & FLOPs (G) & DSC $\uparrow$& Jaccard $\uparrow$& HD95 $\downarrow$& ASD$\downarrow$\\ 
            \midrule
            \midrule 
        UNETR \cite{hatamizadeh2022unetr} & 92.45 & \textbf{63.53} & 77.42 & 63.95 & 15.07 & 5.09 \\
        UNETR++ \cite{shaker2022unetr++} & 96.77 & 102.44 & 80.59 & 68.08 & 8.63 & 2.25 \\
        \midrule
        \rowcolor[HTML]{C8FFFD}
        \textbf{D-LKA Net} & \textbf{62.07} & 166.63 & \textbf{81.22} & \textbf{68.90} & \textbf{7.59} & \textbf{1.99} \\
        \midrule
        \bottomrule
        \end{tabular}
        }
        \vspace{-0.25em}
        
        \label{tab:pancreas_results}
        \end{center}
\vspace{-0.4cm}
\end{table}

\begin{figure}[ht]
    \centering
    \includegraphics[width=\columnwidth]{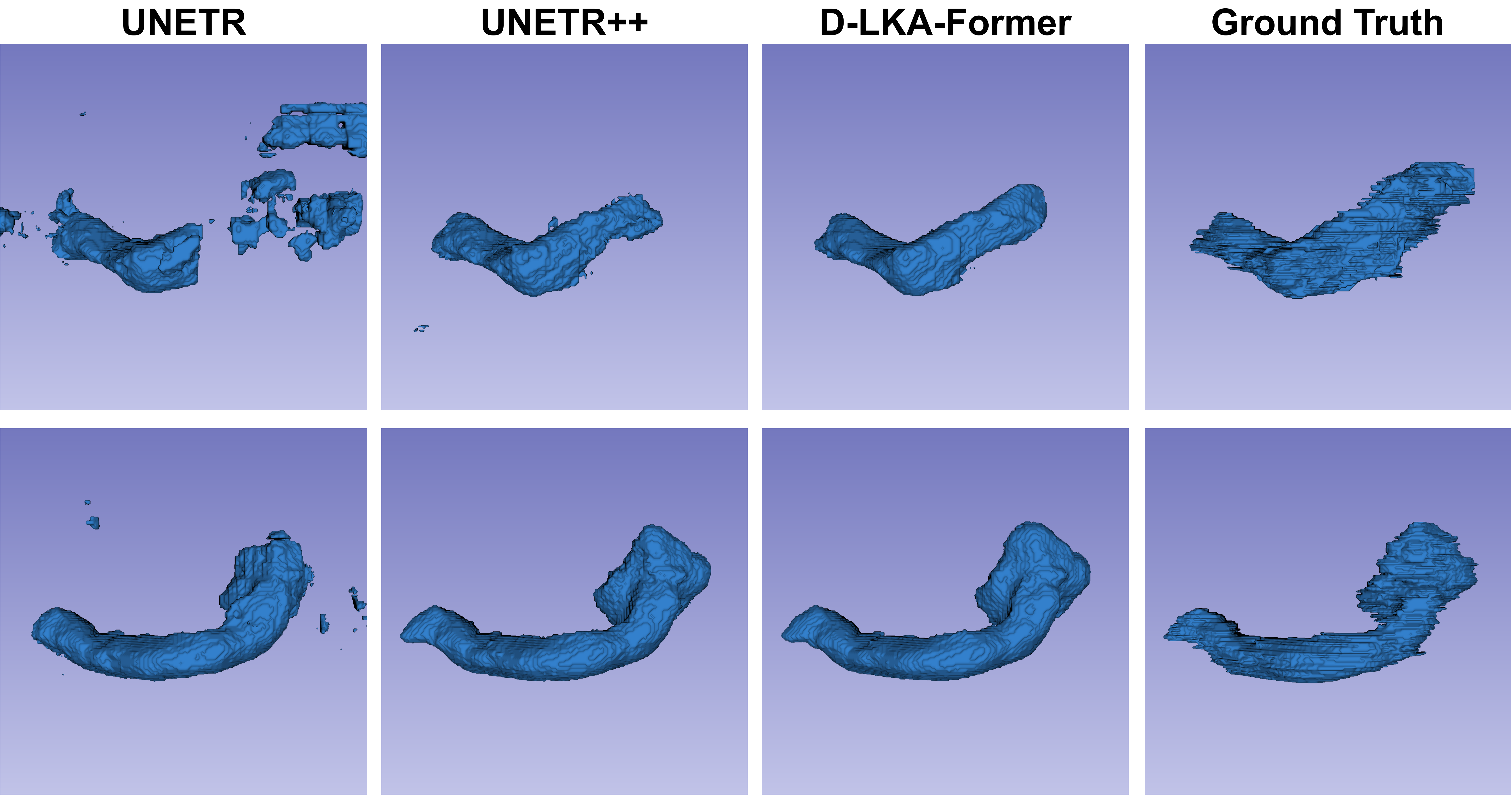}
    \caption{Qualitative results on the Pancreas Dataset.}
    \label{fig:pancreas_3d_visual}
\end{figure}

Figure \ref{fig:pancreas_3d_visual} shows a qualitative comparison of different approaches. UNETR fails to segment the Pancreas as a single object. UNETR++ has smaller artifacts in the segmentation result. Our approach follows the highly irregular shape of the organ better than the other approaches.

\subsection{Ablations Studies}
\noindent\textbf{Robustness.}
To enhance the robustness of our evaluation and to analyze statistical significance, we undertake 5 training runs for each method on the Synapse 2D version. This practice not only ensures a more comprehensive assessment but also enables us to visualize the variations in performance (Please refer to the supplementary file for the visualizations). In our evaluation, we observed a stable increase in performance for the aorta, gallbladder, left and right kidney, liver, pancreas, and stomach, where the median performance is higher than other SOTA methods. Solely the spleen segmentation performance is slightly worse. Furthermore, significant performance improvements are achieved for the gallbladder, pancreas, and stomach.

\noindent\textbf{Deformable LKA Influence.}
We continue our ablation study by determining the effectiveness of D-LKA. For this purpose, we construct a network employing 3D LKA without a deformable layer and another version using 3D LKA with an extra convolutional layer instead of the deformable one. The results of this analysis are presented in Table~\ref{tab:deform_LKA_LKA_LKA3dconv_results}.
Introducing the additional 3D convolutional layer results in a notable improvement in the performance of 0.99\% in DSC when compared to the 3D LKA baseline. However, this modification also increases the number of parameters within the network. Replacing the 3D convolutional layer with a deformable convolutional layer leads to an additional performance boost, as indicated by a 0.63\% increase in DSC. This alteration, similar to the previous one, also introduces more parameters and FLOPs into the network. Since the network's size remains relatively small, the increase in these metrics is acceptable.

\begin{table}[!ht]
    \begin{center}
    \caption{Results for the experiments of the influence of the deformable convolution layer with the Synapse Dataset.}
        \resizebox{\columnwidth}{!}{
        \begin{tabular}{l|c c| c c}
        \toprule
            Methods & Params & FLOPs  & DSC $\uparrow$ &HD95 $\downarrow$  \\
            \midrule
            \midrule
        3D LKA & 28.94 & 48.79 & 85.87 & 14.35  \\
        3D LKA + 3D conv. & 37.73 & 58.64 & 86.86 & 12.69  \\
        3D LKA + 3D deform. conv. & 42.35 & 66.96 & 87.49 & 9.57 \\
        \midrule
        \bottomrule
        \end{tabular}
        }\vspace{0.5em}
        
        \label{tab:deform_LKA_LKA_LKA3dconv_results}
        \end{center}

\end{table}

\noindent\textbf{Skip Connections.}
Lastly, we assess the effect of the skip connections on the segmentation process. The results are shown in Table \ref{tab:deform_LKA_skip_connections_results}. We remove all skip connections and gradually add them to the network, starting with the highest-level skip connection. The results yield that skip connections are crucial for obtaining optimal segmentation performance. Additionally, we highlight that the highest-level skip connection is vital for achieving the best segmentation result, improving the DSC performance by 0.42\%.

\begin{table}[!ht]
    \begin{center}
        \caption{Results for the experiments of the influence of the number of skip connections with the Synapse Dataset.}
        \begin{tabular}{l|c c c c c}
        \toprule
            Metrics & 0 & 1 & 2 & 3 & 4 \\
            \midrule
            \midrule
            DSC $\uparrow$ & 84.78 & 85.99 & 86.56 & 87.07 & 87.49 \\
            \hline
            HD95 $\downarrow$ & 11.18 & 12.21 & 14.89 & 9.25 & 9.57 \\
        \midrule
        \bottomrule
        \end{tabular}
        \vspace{0.5em}
        
        \label{tab:deform_LKA_skip_connections_results}
        \end{center}
\vspace{-0.4cm}
\end{table}

\section{Conclusion}
In this paper, we propose a novel hierarchical hybrid Vision Transformer and CNN architecture using Deformable Large Kernel Attention (D-LKA Net). This attention mechanism enables the network to learn a deformation grid for accessing more relevant information than conventional attention strategies. Furthermore, the Large Kernel Attention mechanism can aggregate global information similar to that of self-attention to overcome the local restrictions of CNN mechanisms. Further, we present a 3D version of our proposed network, which includes cross-slice feature extraction for an even stronger representation capability. Our models receive SOTA results on several publicly available segmentation datasets. Overall, we believe that the proposed D-LKA Net is a robust and powerful choice for medical image segmentation.

{\small
\bibliographystyle{ieee_fullname}
\bibliography{egbib}
}

\clearpage
\newpage
\appendix
\newpage

\section*{Supplementary Material}
The additional materials provided encompass an extended ablation study, which serves to demonstrate the robustness and efficacy of our method in addressing semantic segmentation tasks. Furthermore, we present supplementary visualizations and discussions that accentuate the impactful role played by the D-LKA module that we have proposed.

\section{Computationl complexity of the D-LKA}
A comparison of the number of parameters for normal convolution and the constructed convolution is shown in table \ref{tab:convparametercomparison}. While the numbers of the standard convolution explode for a larger number of channels, the parameters for decomposed convolution are lower in general and do not increase as fast. Deformable decomposed convolution adds a lot of parameters in comparison to decomposed convolution but is still significantly smaller than standard convolution. The main amount of parameters for deformable convolution is created by the offset network. Here, we assumed a kernel size of (5,5) for the deformable depth-wise convolution and (7,7) for the deformable depth-wise dilated convolution. This results in the optimal number of parameters for a large kernel of size $21\times 21$. A more efficient way to generate the offsets would greatly reduce the number of parameters.

\begin{table*}[t]
    \centering
    \caption{The number of parameters for standard convolution and decomposed convolution. The kernel size is $21 \times 21$. Adapted from \cite{guo2022visual}.}
    \begin{tabular}{|c|c|c|c|c|c|}
        \hline
        \# Channels & Std. Conv. & Decomp. Conv. & Deform. Decom. Conv. & Offset DDW-Conv. & Offset DDW-D Conv.\\
        \hline
        $C=32$ & $451,584$  & $3,392$ & $197,204$ & $40,050$ & $153,762$\\
        $C=64$ & $1,806336$ & $8,832$ & $396,308$ & $80,050$ & $307,426$\\
        $C=128$ & $7,225344$ & $25,856$ & $800,660$ & $160,050$ & $614,754$\\
        $C=256$ & $28,901,376$ & $84,480$ & $1,633,940$ & $320,050$ & $1,229,410$\\
        $C=512$ & $115,605,504$ & $300,032$ & $3,398,804$ & $640,050$ & $2,458,722$\\
        \hline
    \end{tabular}
    
    \label{tab:convparametercomparison}
\end{table*}

\begin{figure}[!h]
    \centering
    \includegraphics[width=\columnwidth,keepaspectratio]{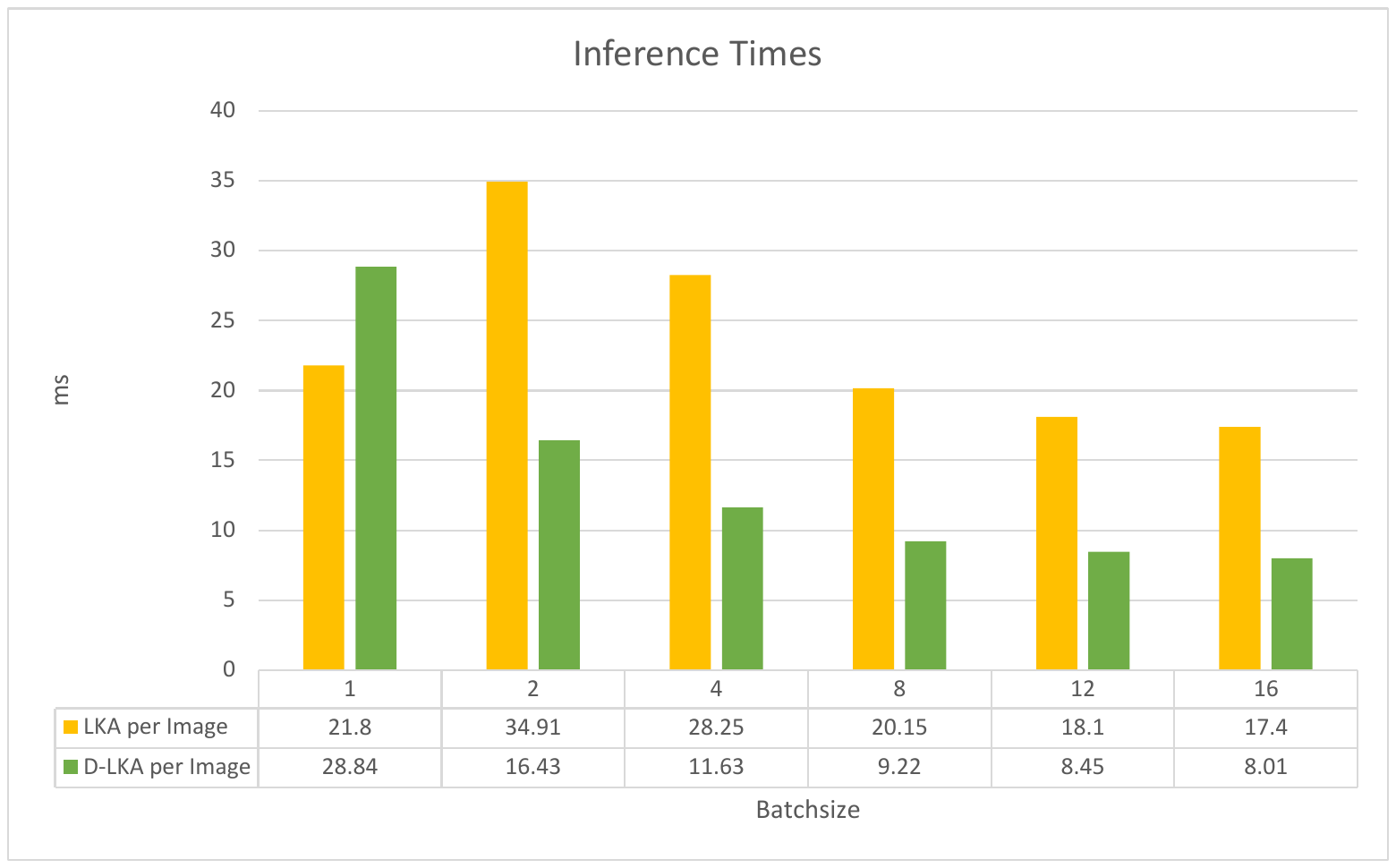}
    \caption{The inferences time in ms for an input image of size $3\times 224 \times 224$ on the 2D methods. The times are already calculated for a single image for better comparison.}
    \label{fig:inferencespeedcomparison}
\end{figure}

It is worth noting that the introduction of the deformable LKA does indeed introduce additional parameters and floating-point operations per second (FLOPS) to the model. However, it's important to emphasize that this increase in computational load does not impact the overall inference speed of our model. Instead, for batch sizes $>1$, we even observe a reduction in inference time, shown in Figure \ref{fig:inferencespeedcomparison}. For instance, based on our extensive experiments, we have observed that for a batch size of 16, the inference times with and without deformable convolution are only 8.01ms and 17.38ms, respectively. We argue that this is due to the efficient implementation of the deformable convolution in 2D. To measure the times, a random input of size $(b \times 3 \times 224 \times 224)$ is used. The network is inferred 1000 times after a GPU warm-up period of 50 iterations. The measurements are done on an NVIDIA RTX 3090 GPU.

\section{Performance vs Efficiency}
\begin{figure}[!h]
    \centering
    \includegraphics[width=\columnwidth,keepaspectratio]{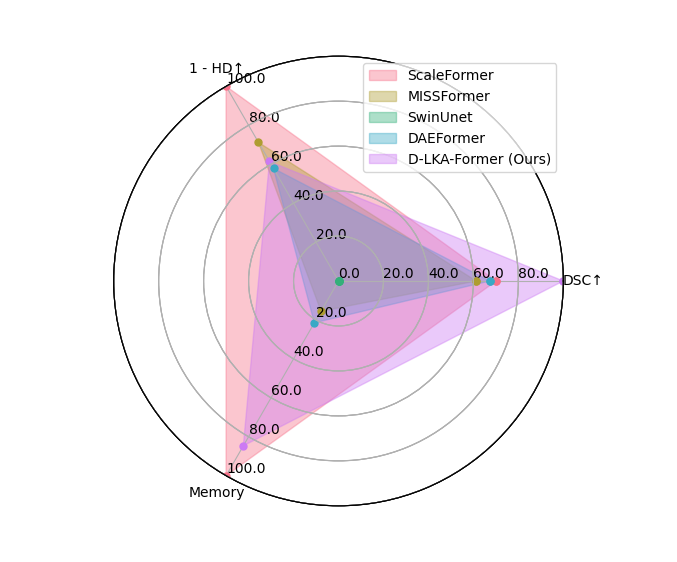}
    \caption{Performance vs memory chart to compare the performance of SOTA approaches, including ScaleFormer \cite{huang2022scaleformer}, MISSFormer \cite{huang2022missformer}, SwinUnet \cite{cao2021swinunet}, DAEFormer \cite{azad2022dae}, with our the proposed 2D D-LKA Net on Synapse dataset. DSC, HD, and Memory values are normalized using min-max normalization for improved visibility and comparability.}
    \label{fig:radarplot}
\end{figure}

To leverage the performance vs. parameter tradeoff, we visualize the performances on the Synapse 2D dataset, reported in DSC and HD, and the memory consumption based on the number of parameters in Figure \ref{fig:radarplot}.
The D-LKA Net induces a rather large amount of parameters with approximately 101M. This is less than the second best-performing method, the ScaleFormer \cite{huang2022scaleformer}, which used 111.6M parameters. Compared to the more light-weight DAEFormer \cite{azad2022dae} model, we, however, achieve a better performance justifying the parameter increase. The majority of the parameters are from the MaxViT encoder; thus, replacing the encoder with a more efficient one can reduce the model parameters.
It's also worth noting that in this visualization, we initially normalized both the HD and memory values within the [0, 100] range. Subsequently, we scaled them down from 100 to enhance the representation of higher values.

\section{Qualitative results on the Synapse dataset}
\begin{figure*}[!ht]
    \centering
    \includegraphics[width=\textwidth]{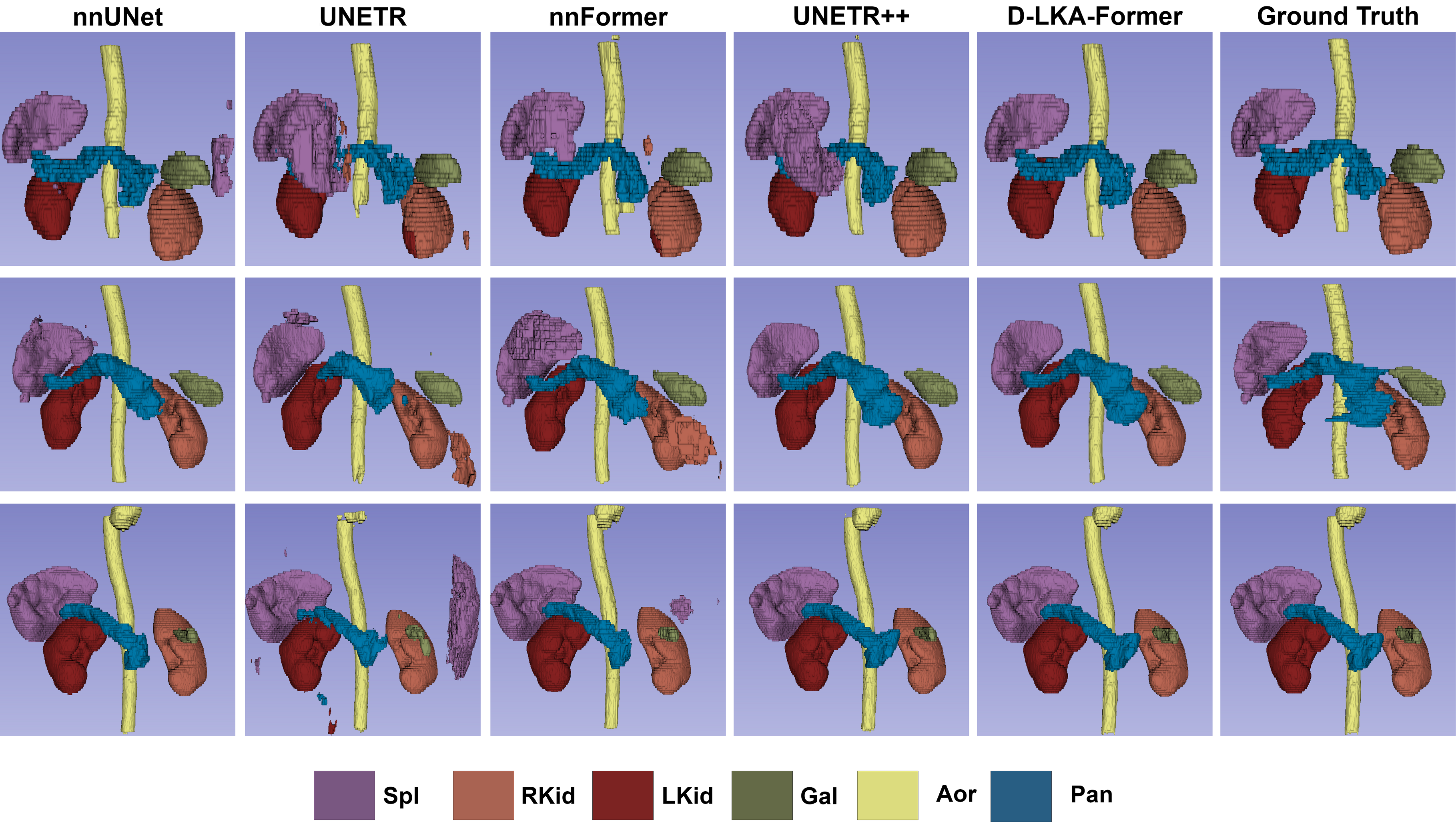}
    \caption{Additional qualitative results on the Synapse Dataset. Liver and Stomach are not shown for improved visibility of smaller occluded organs.}
    \label{fig:synapse_3d_visual_appendix}
\end{figure*}

\begin{figure*}[!ht]
    \centering
    \includegraphics[width=\textwidth]{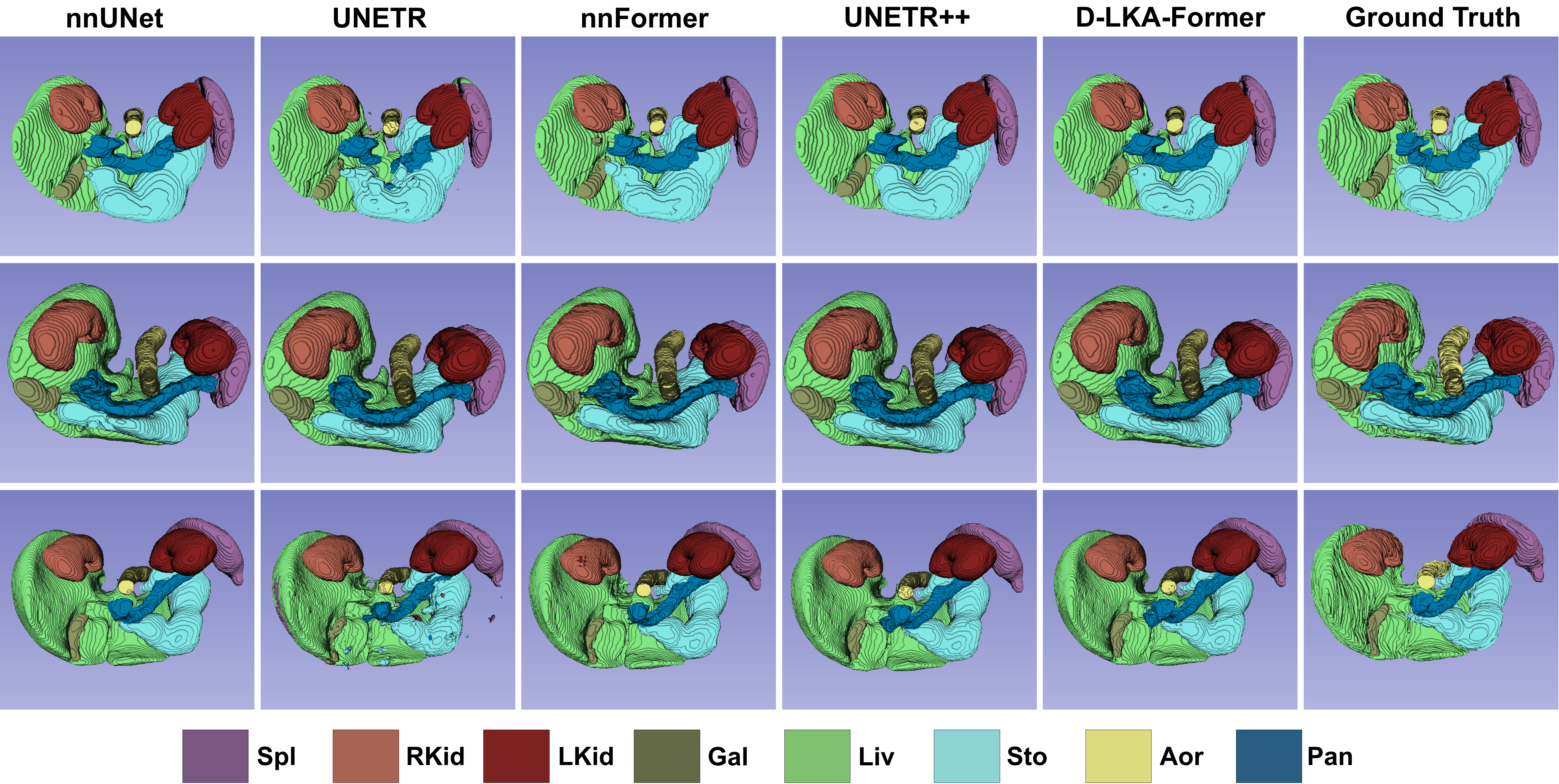}
    \caption{Additional qualitative results on the Synapse Dataset.}
    \label{fig:synapse_3d_visual_appendix_unten}
\end{figure*}

To further visualize our model capability, we provide a different perspective of the 3D organ segmentation of the Synapse dataset in Figure \ref{fig:synapse_3d_visual_appendix} and Figure \ref{fig:synapse_3d_visual_appendix_unten}. We neglect the visualization of the liver and the stomach so partly occluded organs get a better visibility.

\begin{figure*}[!ht]
    \centering
    \resizebox{\textwidth}{0.9\textheight}{
    \includegraphics[width=\textwidth]{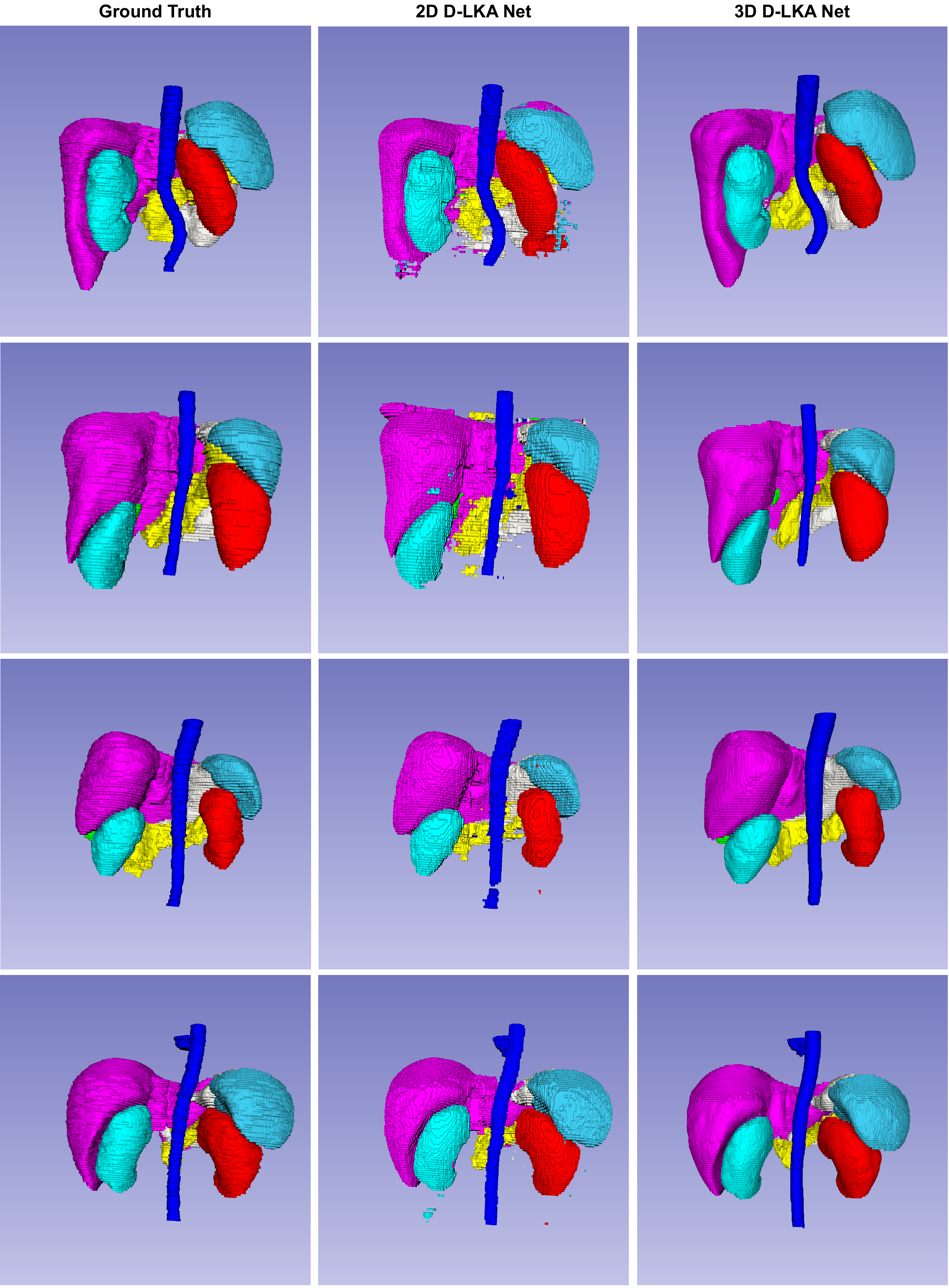}}
    \caption{Additional qualitative results of the 2D D-LKA-Former on the Synapse dataset, visualized in 3D. The comparison to the 3D D-LKA Net is shown. Here, it is visible that the 3D version creates less noise due to the inter-slice dependencies.}
    \label{fig:synapse_2d_in_3d_visual_appendix}
\end{figure*}

To gain a better understanding of the limitations associated with the 2D approach, it is advisable to expand our perspective into the 3D domain. As illustrated in Figure \ref{fig:synapse_2d_in_3d_visual_appendix}, we can observe inconsistencies among the slices. These discrepancies can be attributed to the absence of information exchange between neighboring slices in a 2D network. In contrast, our 3D network successfully mitigates these limitations.

\section{Limitations on the Skin dataset}
\begin{figure*}[!ht]
    \centering
    \includegraphics[width=\textwidth]{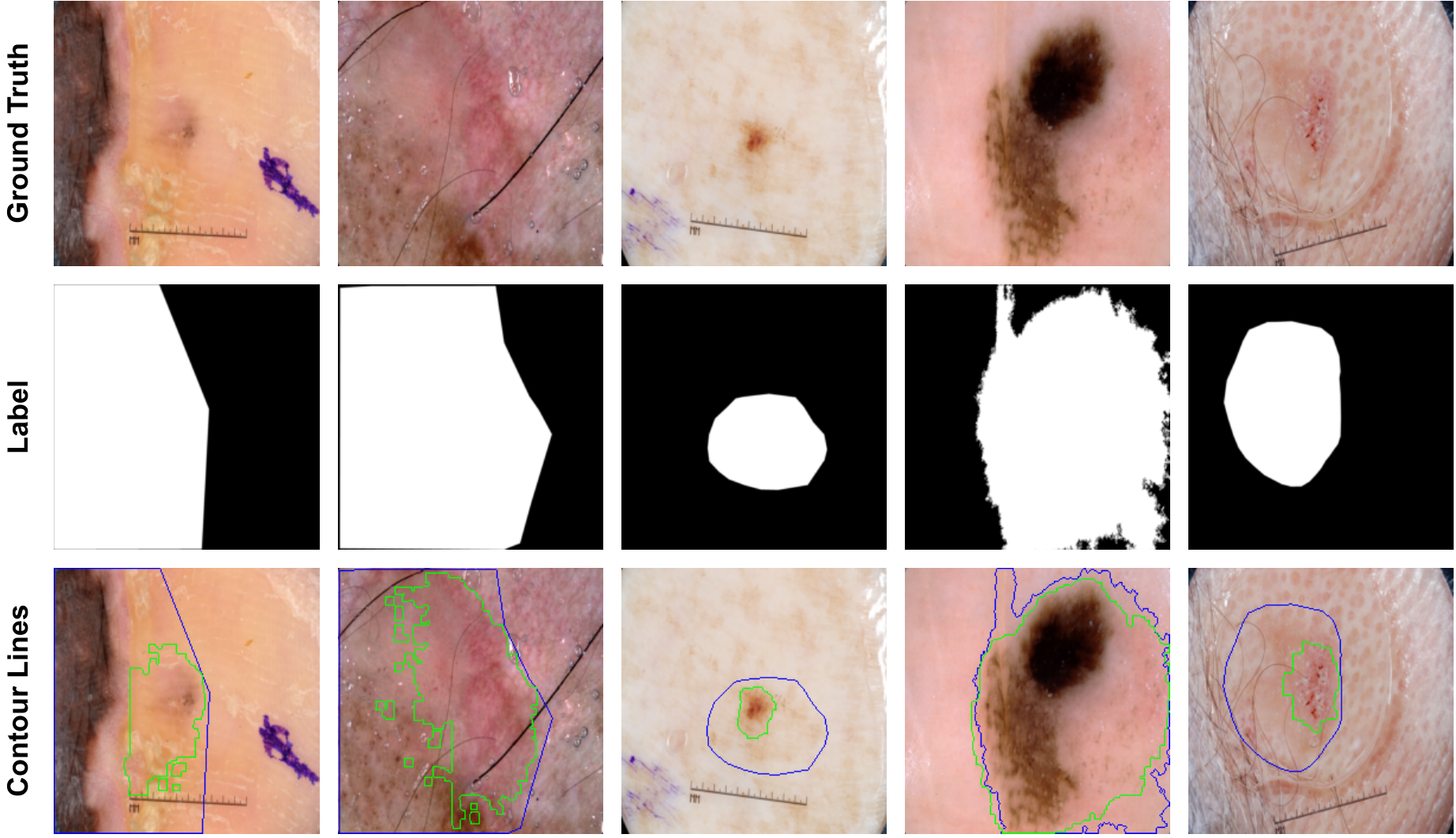}
    \caption{Additional qualitative results of the 2D D-LKA-Former on the ISIC 2018 dataset.}
    \label{fig:skin_visual_appendix}
\end{figure*}

Figure \ref{fig:skin_visual_appendix} shows a qualitative visualization of ISIC 2018 samples, where our approach fails. However, it is also visible that the segmentation is either noisy or quite primitive. Since this is also present in the training data, this could hinder the network from learning accurate segmentations.

\section{Robustness Visualization}
\begin{figure*}[!h]
    \centering
    \includegraphics[width=\textwidth,keepaspectratio]{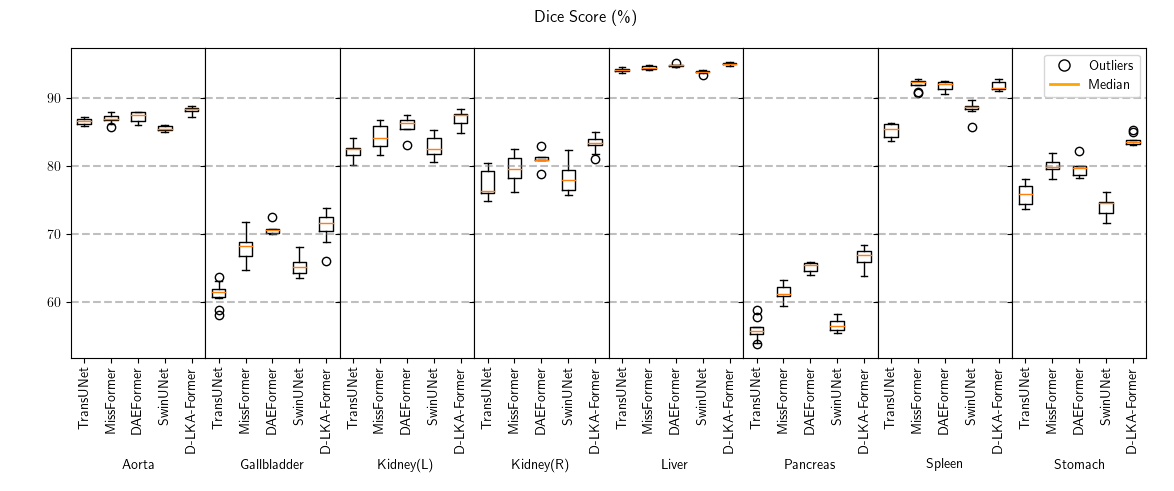}
    \caption{Statistical evaluation of single organ performance of Synapse dataset comparing state-of-the-art methods, including TransUnet \cite{chen2021transunet}, MISSFormer \cite{huang2022missformer}, SwinUnet \cite{cao2021swinunet}, DAEFormer \cite{azad2022dae}, with our the proposed 2D D-LKA Net. Visualized are the results on the Synapse dataset with the performance reported in DSC.}
    \label{fig:boxplot}
\end{figure*}

In line with the ablation study presented in the main paper, we conducted a thorough evaluation of various methods on the Synapse 2D dataset. To ensure the robustness of our findings, we executed each model five times and reported their statistical significance. This detailed analysis is visually represented in Figure \ref{fig:boxplot}.

\end{document}